\documentclass[sigconf]{acmart}
\usepackage[utf8]{inputenc}
\usepackage{xspace}

\usepackage{graphicx}
\usepackage{tikz}
\usepackage{pgfplots}
\usetikzlibrary{pgfplots.groupplots}
\usepgfplotslibrary{fillbetween}
\usepackage{color}
\usepackage{pgfplotstable}
\usetikzlibrary{arrows,petri,topaths,backgrounds,snakes,patterns,positioning}
\pgfplotsset{compat=1.17}
\usepackage{subcaption}
    \usetikzlibrary{
        pgfplots.colorbrewer,
    }
    \pgfplotsset{
cycle list/.define={my marks}{
            every mark/.append style={solid,fill=\pgfkeysvalueof{/pgfplots/mark list fill}},mark=*\\
            every mark/.append style={solid,fill=\pgfkeysvalueof{/pgfplots/mark list fill}},mark=square*\\
            every mark/.append style={solid,fill=\pgfkeysvalueof{/pgfplots/mark list fill}},mark=triangle*\\
            every mark/.append style={solid,fill=\pgfkeysvalueof{/pgfplots/mark list fill}},mark=diamond*\\
        },
    }

\title{Synthetic Graph Generation to Benchmark Graph Learning}
\author{Anton Tsitsulin}
\affiliation{\institution{University of Bonn}
  \country{Germany}
}
\author{Benedek Rozemberczki}
\affiliation{\institution{The University of Edinburgh}
  \country{United Kingdom} }
\author{John Palowitch}
\affiliation{\institution{Google Research}
  \country{USA}
}
\author{Bryan Perozzi}
\affiliation{\institution{Google Research}
  \country{USA}
} \setcopyright{rightsretained}
\copyrightyear{2021}
\acmYear{2021}

\acmConference[GLB'21]{Workshop on Graph Learning Benchmarks at TheWebConf'21}{April 25--29, 2021}{Ljubljana, Slovenia}
\acmBooktitle{GLB'21: Workshop on Graph Learning Benchmarks at TheWebConf, 2021}

\newcommand{\spara}[1]{\smallskip\noindent{\bf #1}}

\definecolor{cycle1}{RGB}{235,172,35}
\definecolor{cycle2}{RGB}{184,0,88}
\definecolor{cycle3}{RGB}{0,140,249}
\definecolor{cycle4}{RGB}{0,110,0}
\definecolor{cycle5}{RGB}{0,187,173}
\definecolor{cycle6}{RGB}{209,99,230}
\definecolor{cycle7}{RGB}{178,69,2}
\definecolor{cycle8}{RGB}{255,146,135}
\definecolor{cycle9}{RGB}{89,84,214}
\definecolor{cycle10}{RGB}{0,198,248}
\definecolor{cycle11}{RGB}{135,133,0}
\definecolor{cycle12}{RGB}{0,167,108}
\definecolor{cyclegray}{RGB}{189,189,189}

\newcommand*{\mD}{\mathbf{D}}
\newcommand*{\mI}{\mathbf{I}}

\newcommand{\ant}[1]{{#1}}
\newcommand{\pal}[1]{{#1}}

\begin{document}

\begin{abstract}
Graph learning algorithms have attained state-of-the-art performance on many graph analysis tasks such as node classification, link prediction, and clustering.
It has, however, become hard to track the field's burgeoning progress.
One reason is due to the very small number of datasets used in practice to benchmark the performance of graph learning algorithms.
This shockingly small sample size (\textasciitilde10) allows for only limited scientific insight into the problem.

In this work, we aim to address this deficiency.
We propose to \emph{generate} synthetic graphs, and study the behaviour of graph learning algorithms in a controlled scenario.
We develop a fully-featured synthetic graph generator that allows deep inspection of different models.
We argue that synthetic graph generations allows for thorough investigation of algorithms and provides more insights than overfitting on three citation datasets.
In the case study, we show how our framework provides insight into unsupervised and supervised graph neural network models.

\end{abstract} \maketitle
\section{Introduction}

In recent years, there has been a tremendous increase in popularity of machine learning on graph-structured data \cite{chami2020machine}.
Most notably, an overwhelming number of Graph Neural Network (GNN) models have been proposed for solving task such as node classification~\cite{kipf2017semi,gat_iclr18}, link prediction~\cite{rozemberczki2021pathfinder}, and graph clustering~\cite{bianchi2020spectral,tsitsulin2020dmon}.
These methods have been applied to several application domains, such as social networks~\cite{deepwalk}, recommender systems~\cite{ying2018graph}, and even molecular analysis~\cite{defferrard2016}.

Despite (or perhaps due to) this abundance of methods, there are still many outstanding issues with benchmarks for graph learning.
In our opinion, the primary issue is that only a few datasets \cite{hu2020open} are actually used when benchmarking algorithms on these tasks.
Making the situation worse, these datasets are remarkably similar to each other; for example, many are derived from from academic citation networks.
Furthermore, some of datasets are particularly ill-suited for use as GNN evaluation, exhibiting extreme structural characteristics, such as abnormally high homophily \cite{shchur2018pitfalls}, and lacking rich features representative of real-world graphs \cite{halcrow2020grale}. We see all this as a critical problem for the field, as the lack of a diverse evaluation suite can give only the illusion of progress.

In this work, we study the use of synthetic graph generation to ameliorate these weaknesses of current graph learning benchmarking.
We propose a fully-featured synthetic graph generator that allows deep inspection of different models.
Synthetic graph generations allows for thorough investigation of algorithms and provides more insights than overfitting on a few citation datasets.
In the case study, we show how our framework provides insight into unsupervised and supervised graph neural network models.

Specifically, our contributions are the following:\begin{itemize}
    \item We unify the ongoing efforts in synthetic evaluation of graph learning algorithms.
    \item We propose a framework that presents a clear path towards new controllable experiments.
    \item We present two case studies that validate the effectiveness of our approach.
\end{itemize}
\vspace{-1ex} \section{Related Work}

We now briefly review recent advances in evaluation of graph learning algorithms.
We describe two main categories of papers: efforts in non-synthetic evaluation, such as establishment of new evaluation protocols and new datasets, and synthetic datasets or benchmark studies.
While there are some contemporary efforts in creating artificial graphs with node features~\cite{cabam2020shah}, no studies have proved their usefulness for benchmarking graph algorithms.

\subsection{Non-Synthetic Benchmark Sets}

While benchmark sets are not the primary focus of this paper, they provide insight into best methodological practices for benchmarking.
Many such best practices, such as fixed train and test split, are not specific to the consideration of synthetic graph data.

TUBenchmark~\cite{morris2020tudataset} is collection of graph classification datasets. 
It includes several biological and chemical graph tasks, primarily on graph classification.
While the feature dimension of this dataset collection is rich---there are both node and edge features present---the sizes of most datasets are very small both in terms of the number of graphs and number of nodes in such graphs. This prevents us from making  meaningful predictions about methods' performance on out-of-domain datasets.
Moreover, there is no guidance for the data split, or metrics used for evaluation for these datasets, limiting the comparability of different methods.

Open Graph Benchmark~\cite{hu2020open} fixes this problem: it unifies the evaluation protocol in terms of the data splits, metrics, and tasks on several datasets representing different application domains and tasks.
Nevertheless, the scarcity of datasets per task is preventing deeper insights into performance of graph learning algorithms.

The first benchmark study to introduce a synthetic task is in~\cite{dwivedi2020benchmarking}. It considers two recipes: semi-supervised graph clustering akin to our proposal from Section~\ref{sec:case-2-semisupervised} and a subgraph search task.
Importantly, node features in the paper are not useful for the task, as they are sampled from a discrete uniform distribution.

\subsection{Synthetic Graph Mining Tasks}

We now review different synthetic generators tailored to generate data for a particular task.
In several graph mining papers, synthetically generated tasks are used to exemplify a particular sensitivity or a feature of an algorithm.
All of them, however, do not leverage the full power of synthetic graph generation, which, as we will exemplify later in the paper, is able to provide insight into many different aspects of graph learning algorithms.

We distinguish three main lines of investigation: unsupervised and semi-supervised clustering tasks, subgraph and substructure search, and learning or varying graph statistics.

\textbf{Graph clustering}~\cite{schaeffer2007graph} is one of the fundamental problems in graph analysis.
In many graph mining tasks, ground-truth labels are highly correlated with the meso- and macroscopic structure of a graph.
The most popular model to generate graphs with a predefined cluster structure is the Stochastic Block Model (SBM)~\cite{snijders1997estimation}.
SBMs generate edges of graphs conditioned on some predefined partition of a graph.
As a general model, many modifications have been proposed in the literature for including additional properties to generated graphs~\cite{lee2019review}, making them particularly suited for the needs of benchmarking.
Stochastic blockmodels have been explored for both unsupervised~\cite{focused,tsitsulin2020dmon} and semi-supervised~\cite{rozemberczki2021pathfinder,dwivedi2020benchmarking} settings.
In the semi-supervised setting, we provide the algorithms additional node labels for a small subset of nodes.
In the extreme case, we label just one node per cluster, as done in~\cite{dwivedi2020benchmarking}.
The task is then to recover latent partitions that generated the data.

\textbf{Substructure search} tasks aim to either find or count some small graph structures embedded in some larger graph.
In this line of work,~\cite{chen2020can} studies how graph neural networks can count motifs (3--4 node subgraphs) in graphs.
In a similar direction,~\cite{dwivedi2020benchmarking} plants a pre-defined graph into random SBM graphs to test the ability of graph neural networks to find exact substructures.

Last, synthetic graphs can help to study how different \textbf{statistical properties} are captures in different graph models.
For instance,~\cite{you2020design} learns statistics such as PageRank or average path length of a graph in order to select well-performing architectures to be transferred to real-world applications.
Synthetic data generation also aids the study of variables that are very hard to control---\cite{zhu2020generalizing} generates random graphs and inserts node features from real ones to control for the degree of homophily in graphs.

 \section{Evaluating Learning Algorithms with Synthetic Graphs}\label{sec:method}

We now describe our proposed methodology for benchmarking graph learning algorithms using generated data.
We apply our model to the unsupervised and semi-supervised graph clustering case, however, we believe that our model can be easily extended to tasks like substructure search and statistical property prediction.

We introduce an \emph{attributed}, \emph{degree-corrected} stochastic block model (ADC-SBM).
The SBM~\citep{snijders1997estimation} plants a partition of clusters (``blocks'') in a graph, and generates edges via a distribution conditional on that partition.
This model has been used extensively to benchmark graph clustering methods~\citep{fortunato2016community}, and has recently been used for experiments on state-of-the-art \pal{supervised} GNNs~\citep{dwivedi2020benchmarking}.
In our version of the model, node features are also generated, using a multivariate mixture model, with the mixture memberships having some association (or de-assocation) with the cluster memberships.
\ant{We proceed to describing the graph generation and feature generation components of our model.}
The code of our model is publicly available on GitHub\footnote{\url{https://github.com/google-research/google-research/tree/master/graph_embedding/simulations}}.

\spara{ADC-SBM graph generation.} We fix a number of nodes $n$ and a number of clusters $k$, and choose node cluster memberships uniformly-at-random. Define the matrix $\mD_{k\times k}$ where $\mD_{ij}$ is the expected number of edges shared between nodes in clusters $i$ and $j$. We determine $\mD$ by fixing (1) the expected \emph{average} degree of the nodes $d\in \{1,2,\dots, n\}$, and (2) the expected \emph{average} sub-degree $d_{out} \leq d$ of a node to any cluster other than its own.
The difference $d_{in} - d_{out}$, where $d_{in}:=d - d_{out}$, controls the spectral detectability of the clusters~\citep{nadakuditi2012graph}.
Finally, we generate a power-law $n$-vector $\theta$ on the range $[d_{min}, d_{max}]$ with exponent $\alpha > 0$. The ADC-SBM is generated so that the expected degree of node $i$ is proportional to $\theta_i$. Note that $d_{min}, d_{max}$ are arbitrary, though their \emph{difference} increases the extremity of the power law. We use the generated memberships and the generated parameters $\mD$ and $\theta$ as inputs to the degree-corrected SBM function from the graph-tool~\cite{peixoto_graph-tool_2014} package.

\begin{figure*}[!t]
\begin{subfigure}[b]{0.24\linewidth}
\centering
\includegraphics[width=\linewidth]{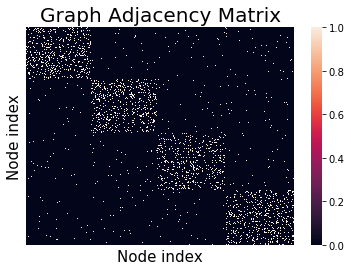}\caption{}
\end{subfigure}\hfill
\begin{subfigure}[b]{0.24\linewidth}
\centering
\includegraphics[width=\linewidth]{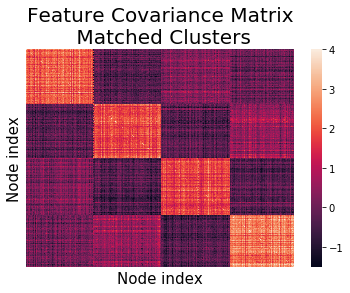}\caption{}
\end{subfigure}\hfill
\begin{subfigure}[b]{.24\linewidth}
\centering
\includegraphics[width=\linewidth]{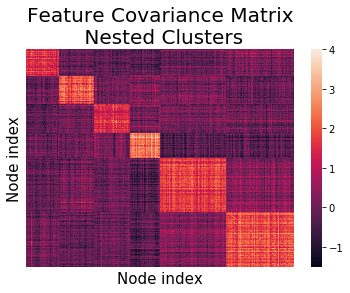}\caption{}
\end{subfigure}\hfill
\begin{subfigure}[b]{0.24\linewidth}
\centering
\includegraphics[width=\linewidth]{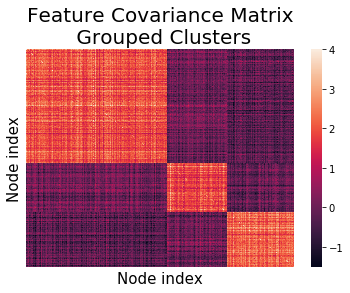}
\caption{}
\end{subfigure}
\caption{\label{fig:synthetic-illustration}\textbf{Illustration of synthetic data.} (a) 4-cluster graph adjacency matrix. (b) Covariance matrix of ``matched'' features: features that are clustered according to the graph clusters. (c) Covariance matrix of ``nested'' features: features that are clustered by incomplete nesting of the graph clusters. (d) Covariance matrix of ``grouped'' features: features that are clustered by incomplete grouping of the graph clusters.}
\end{figure*} 
\spara{ADC-SBM feature generation.} We generate feature memberships from $k_f$ cluster labels. For graph clustering GNNs that operate both on edges and node features, it is important to examine performance on data where feature clusters diverge from or segment the graph clusters: thus potentially $k_f\ne k$. 
Our proposed method affords the creation of feature memberships which \emph{match}, \emph{group}, or \emph{nest} the graph memberships, as illustrated in Figure~\ref{fig:synthetic-illustration}.
With feature memberships in-hand, we generate $k$ zero-mean feature cluster centers from a $s$-multivariate normal with covariance matrix $\sigma_c^2\cdot \mI_{s\times s}$.
Then, for feature cluster $i\leq k_f$, we generate its features from a $s$-multivariate normal with covariance matrix $\sigma^2\cdot \mI_{s\times s}$.
Note that the ratio $\sigma^2_c/\sigma^2$ controls the expected value of the classical between/within-sum-of-squares of the clusters.

We also generate optional edge features in the similar way.
We group edges into intra-class and inter-class ones; we sample intra-class edge features from zero-mean, unit-covariance $s_e$-multivariate normal distribution.
Inter-class edges are sampled with mean $(x_e, x_e, \cdots, x_e)$ and unit covariance.
Essentially, the bigger $x_e$ is, the further apart are inter-class and intra-class edge features, the easier the task of recovering the ground-truth partition is.

 \section{Case studies}

We now describe how an ADC-SBM model can be adopted to study model characteristics in different domains. This is especially useful when there are only a few relevant academic benchmarks available.  First, we examine using the model for studying unsupervised clustering methods using GNNs.
Second, we examine adapting the model to examine the performance of state-of-the-art models for {semi-supervised classification in multigraphs}.

\subsection{Unsupervised Clustering}\label{sec:case-1-clustering}

We begin with \emph{unsupervised} clustering with graph neural network models.
To study model robustness, we define ``default'' ADC-SBM parameters, and explore model parameters in a range around the defaults.
We configure our default model as follows: we generate graphs with $n=1,000$ nodes grouped in $k=4$ clusters, and $s=32$-dimensional features grouped in $k_f=4$ matching feature clusters with $\sigma=1$ intra-cluster center variance and $\sigma_c=3$ cluster center variance. We mimic real-world graphs' degree distribution with $d=20$ average degree and $d_{out}=2$ average inter-cluster degree with power law parameters $d_{min}=2, d_{max}=4, \alpha=2$.
In total, we consider 4 different scenarios.
We take DMoN~\cite{tsitsulin2020dmon}, a modern GNN clustering method, as an example method to be studied, and compare it to three baselines: SBM-based community detection that uses purely local information, k-means over the feature vectors, and Deep Graph Infomax~\cite{velickovic2018}, a baseline that na\"ively combines two signals.
We normalize the signal strength such that clustering with only features or graph structure yields the same performance.

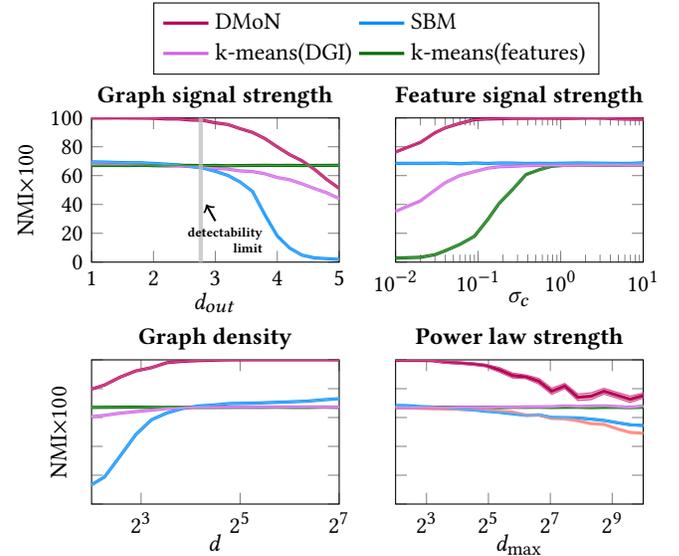
\begin{figure}[tb]
\centering
\begin{tikzpicture}
\begin{groupplot}[group style={
                      group name=myplot,
                      group size= 2 by 2, horizontal sep=0.75cm,vertical sep=1.3cm},height=3.5cm,width=0.575\columnwidth,ymin=0,ymax=100,title style={at={(0.5,0.9)},anchor=south},every axis x label/.style={at={(axis description cs:0.5,-0.15)},anchor=north}]
\nextgroupplot[
 	title = \textbf{Graph signal strength},
 	legend columns=2,
	legend style={at={(1.15,1.3)},anchor=south},
	legend cell align=left,
    legend entries={DMoN, SBM, k-means(DGI), k-means(features)},
	ylabel=NMI$\times100$,
	xlabel=$d_{out}$,
	xmin=1,
	xmax=5
]
\addplot[very thick,color=cycle2] table[x=param,y=mgcn2] {data/scenario1.tex};
\addplot[very thick,color=cycle3] table[x=param,y=sbm] {data/scenario1.tex};
\addplot[very thick,color=cycle6] table[x=param,y=dgi] {data/scenario1.tex};
\addplot[very thick,color=cycle4] table[x=param,y=km] {data/scenario1.tex};

\addplot[name path=mgcn_top_,color=cycle2!70,forget plot] table[x=param,y=mgcn2_high] {data/scenario1.tex};
\addplot[name path=mgcn_btm_,color=cycle2!70,forget plot] table[x=param,y=mgcn2_low] {data/scenario1.tex};
\addplot[cycle2!50,fill opacity=0.5,forget plot] fill between[of=mgcn_top_ and mgcn_btm_];

\addplot[name path=dgi_top_,color=cycle6!70,forget plot] table[x=param,y=dgi_high] {data/scenario1.tex};
\addplot[name path=dgi_btm_,color=cycle6!70,forget plot] table[x=param,y=dgi_low] {data/scenario1.tex};
\addplot[cycle6!50,fill opacity=0.5,forget plot] fill between[of=dgi_top_ and dgi_btm_];

\addplot[name path=sbm_top_,color=cycle3!70,forget plot] table[x=param,y=sbm_high] {data/scenario1.tex};
\addplot[name path=sbm_btm_,color=cycle3!70,forget plot] table[x=param,y=sbm_low] {data/scenario1.tex};
\addplot[cycle3!50,fill opacity=0.5,forget plot] fill between[of=sbm_top_ and sbm_btm_];

\addplot[name path=km_top_,color=cycle4!70,forget plot] table[x=param,y=km_high] {data/scenario1.tex};
\addplot[name path=km_btm_,color=cycle4!70,forget plot] table[x=param,y=km_low] {data/scenario1.tex};
\addplot[cycle4!50,fill opacity=0.5,forget plot] fill between[of=km_top_ and km_btm_];

\draw [color=cyclegray, ultra thick, draw opacity=0.75] (2.76393202250021,0) -- (2.76393202250021,100);
\draw[->,thick,color=black](axis cs:3,28.5)--(axis cs:2.85,40);
\node[anchor=east] (text) at (axis cs:3.9,20){\tiny \textbf{detectability}};
\node[anchor=east] (text2) at (axis cs:3.9,10){\tiny \textbf{limit}};
\nextgroupplot[
 	title = \textbf{Feature signal strength},
	yticklabels={,,},
	xlabel=$\sigma_c$,
	xmode=log,
	xmin=0.01,
	xmax=10
]
\addplot[very thick,color=cycle2] table[x=param,y=mgcn2] {data/scenario2.tex};
\addplot[very thick,color=cycle3] table[x=param,y=sbm] {data/scenario2.tex};
\addplot[very thick,color=cycle4] table[x=param,y=km] {data/scenario2.tex};
\addplot[very thick,color=cycle6] table[x=param,y=dgi] {data/scenario2.tex};

\addplot[name path=mgcn_top_,color=cycle2!70,forget plot] table[x=param,y=mgcn2_high] {data/scenario2.tex};
\addplot[name path=mgcn_btm_,color=cycle2!70,forget plot] table[x=param,y=mgcn2_low] {data/scenario2.tex};
\addplot[cycle2!50,fill opacity=0.5,forget plot] fill between[of=mgcn_top_ and mgcn_btm_];

\addplot[name path=sbm_top_,color=cycle3!70,forget plot] table[x=param,y=sbm_high] {data/scenario2.tex};
\addplot[name path=sbm_btm_,color=cycle3!70,forget plot] table[x=param,y=sbm_low] {data/scenario2.tex};
\addplot[cycle3!50,fill opacity=0.5,forget plot] fill between[of=sbm_top_ and sbm_btm_];

\addplot[name path=km_top_,color=cycle4!70,forget plot] table[x=param,y=km_high] {data/scenario2.tex};
\addplot[name path=km_btm_,color=cycle4!70,forget plot] table[x=param,y=km_low] {data/scenario2.tex};
\addplot[cycle4!50,fill opacity=0.5,forget plot] fill between[of=km_top_ and km_btm_];

\addplot[name path=dgi_top_,color=cycle6!70,forget plot] table[x=param,y=dgi_high] {data/scenario2.tex};
\addplot[name path=dgi_btm_,color=cycle6!70,forget plot] table[x=param,y=dgi_low] {data/scenario2.tex};
\addplot[cycle6!50,fill opacity=0.5,forget plot] fill between[of=dgi_top_ and dgi_btm_];
\nextgroupplot[
 	title = \textbf{Graph density},
	yticklabels={,,},
	xlabel=$d$,
	xmode=log,
	xmin=4,
	xmax=128,
	ylabel=NMI$\times100$,
	log basis x={2}
]
\addplot[very thick,color=cycle2] table[x=param,y=mgcn2] {data/scenario5.tex};
\addplot[very thick,color=cycle3] table[x=param,y=sbm] {data/scenario5.tex};
\addplot[very thick,color=cycle4] table[x=param,y=km] {data/scenario5.tex};
\addplot[very thick,color=cycle6] table[x=param,y=dgi] {data/scenario5.tex};

\addplot[name path=mgcn_top_,color=cycle2!70,forget plot] table[x=param,y=mgcn2_high] {data/scenario5.tex};
\addplot[name path=mgcn_btm_,color=cycle2!70,forget plot] table[x=param,y=mgcn2_low] {data/scenario5.tex};
\addplot[cycle2!50,fill opacity=0.5,forget plot] fill between[of=mgcn_top_ and mgcn_btm_];

\addplot[name path=sbm_top_,color=cycle3!70,forget plot] table[x=param,y=sbm_high] {data/scenario5.tex};
\addplot[name path=sbm_btm_,color=cycle3!70,forget plot] table[x=param,y=sbm_low] {data/scenario5.tex};
\addplot[cycle3!50,fill opacity=0.5,forget plot] fill between[of=sbm_top_ and sbm_btm_];

\addplot[name path=km_top_,color=cycle4!70,forget plot] table[x=param,y=km_high] {data/scenario5.tex};
\addplot[name path=km_btm_,color=cycle4!70,forget plot] table[x=param,y=km_low] {data/scenario5.tex};
\addplot[cycle4!50,fill opacity=0.5,forget plot] fill between[of=km_top_ and km_btm_];

\addplot[name path=dgi_top_,color=cycle6!70,forget plot] table[x=param,y=dgi_high] {data/scenario5.tex};
\addplot[name path=dgi_btm_,color=cycle6!70,forget plot] table[x=param,y=dgi_low] {data/scenario5.tex};
\addplot[cycle6!50,fill opacity=0.5,forget plot] fill between[of=dgi_top_ and dgi_btm_];
\nextgroupplot[
 	title = \textbf{Power law strength},
	yticklabels={,,},
	xlabel=$d_{\max}$,
	xmode=log,
	xmin=4,
	xmax=1024,
	log basis x={2},
    max space between ticks=20,
]
\addplot[very thick,color=cycle2] table[x=param,y=mgcn2] {data/scenario6.tex};
\addplot[very thick,color=cycle3] table[x=param,y=sbm] {data/scenario6.tex};
\addplot[very thick,color=cycle4] table[x=param,y=km] {data/scenario6.tex};
\addplot[very thick,color=cycle8] table[x=param,y=deepwalk] {data/scenario6.tex};
\addplot[very thick,color=cycle6] table[x=param,y=dgi] {data/scenario6.tex};

\addplot[name path=mgcn_top_,color=cycle2!70,forget plot] table[x=param,y=mgcn2_high] {data/scenario6.tex};
\addplot[name path=mgcn_btm_,color=cycle2!70,forget plot] table[x=param,y=mgcn2_low] {data/scenario6.tex};
\addplot[cycle2!50,fill opacity=0.5,forget plot] fill between[of=mgcn_top_ and mgcn_btm_];

\addplot[name path=sbm_top_,color=cycle3!70,forget plot] table[x=param,y=sbm_high] {data/scenario6.tex};
\addplot[name path=sbm_btm_,color=cycle3!70,forget plot] table[x=param,y=sbm_low] {data/scenario6.tex};
\addplot[cycle3!50,fill opacity=0.5,forget plot] fill between[of=sbm_top_ and sbm_btm_];

\addplot[name path=km_top_,color=cycle4!70,forget plot] table[x=param,y=km_high] {data/scenario6.tex};
\addplot[name path=km_btm_,color=cycle4!70,forget plot] table[x=param,y=km_low] {data/scenario6.tex};
\addplot[cycle4!50,fill opacity=0.5,forget plot] fill between[of=km_top_ and km_btm_];

\addplot[name path=dgi_top_,color=cycle6!70,forget plot] table[x=param,y=dgi_high] {data/scenario6.tex};
\addplot[name path=dgi_btm_,color=cycle6!70,forget plot] table[x=param,y=dgi_low] {data/scenario6.tex};
\addplot[cycle6!50,fill opacity=0.5,forget plot] fill between[of=dgi_top_ and dgi_btm_];
\end{groupplot}
\end{tikzpicture}
\caption{Synthetic results on the ADC-SBM model with 4 different scenarios. Synthetic graph generation allow us to gain insight into sensitivity and robustness of different graph learning algorithms. DMoN leverages information from both graph structure and node attributes while being more robust to extremal changes in graph structure.}\label{fig:synthetic-results-baselines}
\end{figure} 
\subsubsection{Experimental Findings and Discussion.}
\ant{To better understand the limits of the task, we study the performance of our baselines and report the results on Figure~\ref{fig:synthetic-results-baselines}.
In particular, our interest lies in the performance of the SBM and pure k-means over features, as these two baselines depict the performance possible when utilizing only one aspect of the data.

When varying graph signal strength, DMoN can effectively leverage the feature signal to obtain outstanding clustering performance even when the graph structure is close to random, far beyond the spectral detectability threshold (pictured in gray).
As for the feature signal strength, even in the presence of a weak feature signal DMoN outperforms stochastic SBM minimization.
On the other hand, k-means(DGI) offers some improvements over using features or the graph structure alone, but it never surpasses the strongest signal provider in the graph.
Specifically, k-means(DGI) is never better than the best of k-means(features) or SBM.
Graph density and power law strength scenarios serve as an example of robustness studies that are possible with our framework.
For instance, sparse regime with low average degree is notoriously difficult for graph-only SBM, however, for the DMoN, that is able to leverage both the graph and feature signal, the performance is never worse than the k-means(features) baseline.
}

\begin{figure}[!t]
\centering
\begin{tikzpicture}
\begin{groupplot}[group style={
                      group name=myplot,
                      group size= 2 by 3, horizontal sep=0.75cm,vertical sep=1.2cm},height=3.74cm,width=4.8cm, ymin=-10,ymax=110,ytick={0,20,40,60,80,100},title style={at={(0.5,0.9)},anchor=south},every axis x label/.style={at={(axis description cs:0.5,-0.15)},anchor=north},
cycle list/Set2,
mark list fill={.!75!white},
cycle multiindex* list={
                Set2
                    \nextlist
                my marks
                    \nextlist
                [3 of]linestyles
                    \nextlist
                very thick
                    \nextlist
            },]
            
\nextgroupplot[
 	title = \textbf{Graph density},
 	legend columns=4,
	legend style={at={(1.10,1.25)},anchor=south},
    legend entries={PDN, GCN, GAT, APPNP},
	ylabel=Accuracy$\times100$,
	xlabel=$d$,
	xtick={4,12,20,28},
	xticklabels={4,12,20,28},
	ytick={0,25,50,75,100},
	yticklabels={0,25,50,75,100},
	xmin=2,
	xmax=30,
	ymin=0,
]
\addplot+[error bars/.cd, y dir=both, y explicit]coordinates {
(4,72.957) +- (0.809,0.809)
(8,72.043) +- (0.571,0.571)
(12,68.663) +- (1.513,1.513)
(16,68.565) +- (1.03,1.03)
(20,65.283) +- (0.994,0.994)
(24,60.326) +- (1.183,1.183)
(28,60.522) +- (1.129,1.129)

};
\addplot+[error bars/.cd, y dir=both, y explicit]coordinates {
(4,50.076) +- (0.609,0.609)
(8,50.217) +- (0.494,0.494)
(12,49.435) +- (0.306,0.306)
(16,50.75) +- (0.248,0.248)
(20,48.13) +- (0.496,0.496)
(24,51.554) +- (0.638,0.638)
(28,50.902) +- (0.385,0.385)

};
\addplot+[error bars/.cd, y dir=both, y explicit]coordinates {

(4,42.652) +- (0.494,0.494)
(8,41.37) +- (0.399,0.399)
(12,41.087) +- (0.265,0.265)
(16,41.772) +- (0.39,0.39)
(20,39.185) +- (0.427,0.427)
(24,40.054) +- (0.738,0.738)
(28,39.272) +- (0.491,0.491)

};
\addplot+[error bars/.cd, y dir=both, y explicit]coordinates {
(4,60.62) +- (0.608,0.608)
(8,60.011) +- (1.321,1.321)
(12,61.457) +- (1.007,1.007)
(16,61.598) +- (0.859,0.859)
(20,61.652) +- (0.953,0.953)
(24,60.728) +- (0.654,0.654)
(28,59.446) +- (1.533,1.533)

};

\nextgroupplot[
 	title = \textbf{Feature space size},
	yticklabels={,,},
	xlabel=$s$,
	xtick={2,8,32,128},
	ytick={0,25,50,75,100},
	yticklabels={,,},
	xmode=log,
	xmin=1.41,
	xmax=181,
	ymin=0,	
	log basis x={2}
	]
\addplot+[error bars/.cd, y dir=both, y explicit]coordinates {
(2,50.163) +- (1.0,1.0)
(4,66.663) +- (0.998,0.998)
(8,76.054) +- (0.882,0.882)
(16,93.783) +- (0.284,0.284)
(32,98.065) +- (0.086,0.086)
(64,95.554) +- (0.934,0.934)
(128,79.043) +- (1.655,1.655)

};

\addplot+[error bars/.cd, y dir=both, y explicit]coordinates {
(2,41.239) +- (0.503,0.503)
(4,48.674) +- (0.446,0.446)
(8,57.554) +- (0.237,0.237)
(16,62.391) +- (0.423,0.423)
(32,65.815) +- (0.268,0.268)
(64,69.054) +- (0.238,0.238)
(128,69.283) +- (0.223,0.223)

};

\addplot+[error bars/.cd, y dir=both, y explicit]coordinates {
(2,39.413) +- (0.495,0.495)
(4,43.728) +- (0.613,0.613)
(8,40.728) +- (0.576,0.576)
(16,42.935) +- (0.598,0.598)
(32,37.489) +- (0.36,0.36)
(64,34.902) +- (0.393,0.393)
(128,33.848) +- (0.29,0.29)

};

\addplot+[error bars/.cd, y dir=both, y explicit]coordinates {
(2,46.0) +- (1.406,1.406)
(4,59.837) +- (1.33,1.33)
(8,83.707) +- (0.78,0.78)
(16,96.228) +- (0.185,0.185)
(32,99.207) +- (0.045,0.045)
(64,99.913) +- (0.014,0.014)
(128,99.946) +- (0.01,0.01)

};

\nextgroupplot[
 	title = \textbf{Graph signal strength},
  	ylabel=Accuracy$\times100$,
	xlabel=$d_{out}/d$,
	xtick={0,0.2,0.4,0.6,0.8,1.0},
	ytick={0,25,50,75,100},
	yticklabels={0,25,50,75,100},
	xmin=-.05,
	xmax=1.05,
	ymin=0,	
]
 \addplot+[error bars/.cd, y dir=both, y explicit]coordinates {
(0.0,72.72) +- (0.765,0.765)
(0.2,65.92) +- (1.099,1.099)
(0.4,64.04) +- (1.291,1.291)
(0.6,63.62) +- (0.965,0.965)
(0.8,62.02) +- (0.809,0.809)
(1.0,62.06) +- (1.117,1.117)

};
 \addplot+[error bars/.cd, y dir=both, y explicit]coordinates {
(0.0,53.92) +- (0.431,0.431)
(0.2,52.64) +- (0.462,0.462)
(0.4,51.5) +- (0.396,0.396)
(0.6,51.0) +- (0.302,0.302)
(0.8,50.94) +- (0.504,0.504)
(1.0,47.08) +- (0.414,0.414)

};

 \addplot+[error bars/.cd, y dir=both, y explicit]coordinates {
(0.0,50.2) +- (0.518,0.518)
(0.2,48.3) +- (0.301,0.301)
(0.4,48.34) +- (0.308,0.308)
(0.6,48.46) +- (0.451,0.451)
(0.8,47.86) +- (0.559,0.559)
(1.0,46.7) +- (0.65,0.65)

};

 \addplot+[error bars/.cd, y dir=both, y explicit]coordinates {
(0.0,72.3) +- (0.682,0.682)
(0.2,70.54) +- (0.872,0.872)
(0.4,69.5) +- (0.695,0.695)
(0.6,68.68) +- (0.981,0.981)
(0.8,66.3) +- (0.813,0.813)
(1.0,64.84) +- (1.14,1.14)

};

\nextgroupplot[
 	title = \textbf{Feature signal strength},
	yticklabels={,,},
	xlabel=$\sigma_c$,
	ytick={0,25,50,75,100},
	yticklabels={,,},
	xmode=log,
	xtick={0.125,0.5,2,8},	
	log basis x={2},
	xmin=0.088,
	xmax=11.31,
	ymin=0,	
]
 \addplot+[error bars/.cd, y dir=both, y explicit]coordinates {
(0.125,27.326) +- (0.233,0.233)
(0.25,32.033) +- (0.313,0.313)
(0.5,39.902) +- (0.541,0.541)
(1,69.022) +- (1.604,1.604)
(2,83.533) +- (1.406,1.406)
(4,89.141) +- (1.47,1.47)
(8,93.054) +- (1.12,1.12)

};
 \addplot+[error bars/.cd, y dir=both, y explicit]coordinates {
(0.125,26.337) +- (0.143,0.143)
(0.25,28.402) +- (0.288,0.288)
(0.5,37.641) +- (0.345,0.345)
(1,50.065) +- (0.409,0.409)
(2,58.163) +- (0.532,0.532)
(4,67.228) +- (0.464,0.464)
(8,70.848) +- (0.381,0.381)

};

 \addplot+[error bars/.cd, y dir=both, y explicit]coordinates {
(0.125,24.38) +- (0.107,0.107)
(0.25,26.533) +- (0.243,0.243)
(0.5,32.728) +- (0.317,0.317)
(1,39.609) +- (0.481,0.481)
(2,53.011) +- (0.615,0.615)
(4,55.652) +- (0.619,0.619)
(8,58.5) +- (1.012,1.012)

};

 \addplot+[error bars/.cd, y dir=both, y explicit]coordinates {
(0.125,26.022) +- (0.254,0.254)
(0.25,31.457) +- (0.532,0.532)
(0.5,41.728) +- (0.822,0.822)
(1,65.402) +- (0.794,0.794)
(2,80.761) +- (1.016,1.016)
(4,85.696) +- (1.469,1.469)
(8,74.043) +- (2.476,2.476)

};

\end{groupplot}
\end{tikzpicture}
\vspace*{-6mm}
\caption{Semi-supervised node classification results on the ADC-SBM generated synthetic graphs evaluated by average accuracy score (100 experimental runs).}\label{fig:node_classification}
\vspace{-5mm}
\end{figure}
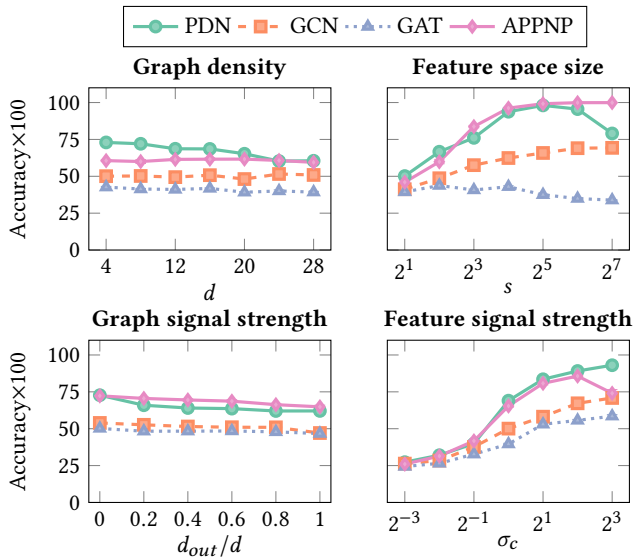

\subsection{Semi-Supervised Node Classification}\label{sec:case-2-semisupervised}
The semi-supervised node classification experiments used a modified set of the default ADC-SBM graph and feature generation hyperparameters from Subsection \ref{sec:case-1-clustering}. We used $s=4$-dimensional node features with $\sigma_c=1$ cluster center standard deviation and the ratio of inter-class was set as $d_{out}/d=0.5$. We generated $s_e=4$ dimensional edge features with a standard deviation of $\sigma_e=0.5$ and the shift of the inter and intra-cluster edge feature distributions was $\Delta \sigma=2$.

Our node classification experiments compared the predictive performance of state-of-the-art graph convolutional models \cite{ppnp_iclr19,bojchevski2020pprgo,gat_iclr18,kipf2017semi,rozemberczki2021pathfinder} using synthetic graphs generated with the default settings of ADC-SBM. In each experiment one of the graph generation hyperparameters was modulated and we evaluated the semi-supervised classifiers with $20$-shot learning tasks \cite{shchur2018pitfalls} and plotted the mean test accuracy scores on Figure \ref{fig:node_classification}. Each model was trained for 200 epochs with a learning rate of $10^{-2}$ and had $2^5$ dimensional convolutional (and edge aggregation) filters. 

\subsubsection{Experimental Findings and Discussion.} Our results in Scenario 1 show that models (GCN, GAT, APPNP) which do not utilize edge features are robust to graph densification as their predictive performance is unaffected. Increasing the dimensionality of the feature space in Scenario 2 initially helps the predictive performance, but the models start to overfit. The results in Scenario 3 and 4 demonstrate that the performance of models depends more on the quality of the node feature signal, than the graph signal -- namely the intra-cluster density of the graph. 

 \section{Conclusion}

This paper introduces the first concise set of evaluation protocols to study and evaluate graph learning algorithms on generated graphs.
We unify the ongoing efforts in model evaluation and propose a unifying framework with a clear path towards scientific investigation of graph learning algorithms.
We verify that our framework delivers actionable scientific insights on two case studies for unsupervised and supervised graph learning models. \bibliographystyle{ACM-Reference-Format}
\bibliography{main}
\end{document}